\documentclass[final,5p,times,twocolumn]{ourarticle}

\usepackage{algorithm}
\usepackage{algorithmic}
\usepackage{amsfonts}
\usepackage{amsmath}
\usepackage{amssymb}
\usepackage{amsthm}
\usepackage{bm}
\usepackage{color}
\usepackage{enumitem}
\usepackage{epstopdf}
\usepackage{graphicx}
\usepackage{mathrsfs}
\usepackage{natbib}
\usepackage{subfigure} 
\usepackage{tikz}
\usepackage{times}

\usetikzlibrary{arrows,calc,shapes,snakes,automata,backgrounds,fit,petri}

\allowdisplaybreaks[1]

\graphicspath{{figures/}}

\def\chol{\mathop{\sf chol}}

\def\diag{\mathsf{diag}}
\def\prior{\mathrm{prior}}
\def\tr{\mathop{\sf tr}}

\def\vec{\mathsf{vec}}

\def\blambda{\bm{\lambda}}
\def\bLambda{\mathbf{\Lambda}}
\def\bmu{\bm{\mu}}

\def\bSigma{\mathbf{\Sigma}}

\def\bxi{\bm{\xi}}

\def\0{\mathbf{0}}
\def\1{\mathbf{1}}

\def\g{\mathbf{g}}
\def\f{\mathbf{f}}

\def\p{\mathbf{p}}
\def\r{\mathbf{r}}
\def\s{\mathbf{s}}
\def\t{\mathbf{t}}

\def\w{\mathbf{w}}
\def\x{\mathbf{x}}
\def\y{\mathbf{y}}
\def\z{\mathbf{z}}
\def\A{\mathbf{A}}
\def\B{\mathbf{B}}

\def\G{\mathbf{G}}

\def\I{\mathbf{I}}

\def\K{\mathbf{K}}
\def\L{\mathbf{L}}
\def\M{\mathbf{M}}

\def\S{\mathbf{S}}
\def\T{\mathbf{T}}

\def\V{\mathbf{V}}

\def\drm{\mathrm{d}}
\def\erm{\mathrm{e}} 

\def\Lcal{\mathcal{L}}

\def\Ncal{\mathcal{N}}

\def\Ical{\mathcal{I}}

\def\Rbb{\mathbb{R}}

\def\old{\mathsf{old}}

\def\defined{\stackrel{.}{=}}
\newcommand{\appropto}{\mathrel{\vcenter{
  \offinterlineskip\halign{\hfil$##$\cr
    \propto\cr\noalign{\kern2pt}\sim\cr\noalign{\kern-2pt}}}}}

\definecolor{Gray}{rgb}{0.4,0.4,0.4}
\definecolor{Blue}{rgb}{0.0,0.0,1.0}
\definecolor{Green}{rgb}{0.07,0.72,0.16}
\definecolor{Purple}{rgb}{0.63,0.1,0.76}

\newcommand{\greentext}[1]{{\color{Green}{#1}}}
\newcommand{\purpletext}[1]{{\color{Purple}{#1}}}

\newcommand{\algofunc}[1]{{\purpletext{\textbf{#1}}}}

\usepackage[pscoord]{eso-pic}

\newcommand{\placetextbox}[3]{
  \setbox0=\hbox{#3}
  \AddToShipoutPictureFG*{
    \put(\LenToUnit{#1\paperwidth},\LenToUnit{#2\paperheight}){\vtop{{\null}\makebox[0pt][c]{#3}}}%
  }%
}%

\begin{document}

\placetextbox{0.5}{0.99}{\small \color{Gray}{\emph{Technical report accompanying ``An Adaptive Resample-Move Algorithm for Estimating Normalizing Constants'' (2016)
\cite{fraccaro16adaptive} }}}%

\begin{frontmatter}

\title{An Efficient Implementation of Riemannian Manifold Hamiltonian Monte Carlo
for Gaussian Process Models}

\author{Ulrich Paquet\corref{corresponding}}
\ead{ulrich@cantab.net}

\author[dtu]{Marco Fraccaro}
\ead{marfra@dtu.dk}

\cortext[corresponding]{Corresponding author}
\address[dtu]{Technical University of Denmark, Lyngby, Denmark}


\begin{abstract} 
This technical report accompanies \cite{fraccaro16adaptive} and
presents pseudo-code for a Riemannian manifold Hamiltonian Monte Carlo (RMHMC) method
to efficiently simulate samples from $N$-dimensional posterior distributions $p(\x | \y)$, where $\x \in \Rbb^{N}$
is drawn from a Gaussian Process (GP) prior, and observations $y_n$ are independent given $x_n$.
Sufficient technical and algorithmic details are provided for the implementation of RMHMC for distributions arising from GP priors.
\end{abstract} 

\end{frontmatter}

\section{Introduction}\label{sec:intro}

When data is modelled with Gaussian process (GP) priors, the resulting posterior distributions are usually highly correlated.
There are various avenues to simulating samples from such posterior distributions with Markov chain Monte Carlo (MCMC) methods,
ranging from simple
Metropolis-Hastings (MH) methods with symmetric proposal distributions to
component-wise Gibbs samplers, to 
more advanced Hamiltonian Monte Carlo (HMC) methods that use the gradient of the log-posterior to guide the sampler
towards high-density regions.
The strong posterior correlations can adversely affect the mixing rates of these methods.

The mixing rates of an MCMC method can be increased significantly when, instead of using only first-order gradient information,
one additionally relies on local second-order statistics of the log-posterior to guide the sampler.
In this note, 
we present \textbf{pseudo-code} for a \textbf{Riemannian manifold Hamiltonian Monte Carlo} (RMHMC) method
\cite{RMHMC}
to efficiently simulate samples from $N$-dimensional posterior distributions
$p(\x | \y)$, where $\x \in \Rbb^N$ is drawn from a zero-mean GP prior
\begin{equation} \label{eq:gpprior}
\prior(\x) = \Ncal(\x ; \0, \K)
\end{equation}
with kernel matrix $\K$.
The aim of this note is to provide sufficient technical and algorithmic details for anyone to implement RMHMC for GPs. 
We assume that the likelihood for each observation $y_n$ depends only on the latent function value $x_n$ through
\begin{equation} \label{eq:loglikelihood}
\ell_n(x_n) \defined \log p(y_n | x_n)
\end{equation}
and is not Gaussian. The resulting posterior or target distribution is
\begin{equation} \label{eq:posterior}
p(\x | \y) = \frac{\prod_{n} p(y_n | x_n) \cdot \prior(\x)}{p(\y)} \ .
\end{equation}
The variables in (\ref{eq:posterior}) often form a very correlated high dimensional density,
which is ideally suited to RMHMC.
One could also sample and infer the GP kernel hyperparameters that govern $\K$.
Additionally sampling kernel hyperparameters is outside the scope of this note,
but a sketch for how they are sampled
for a fully Gaussian model is given in \cite{girolami09riemannian}.

\subsection{A formulation for obtaining normalising constants}

Aside from sampling from $p(\x|\y)$, we are also interested in using samples to estimate $\log Z \defined \log p(\y)$.
We will write (\ref{eq:posterior}) in a slightly more general form,
so that the normalising constant $Z \defined p(\y)$ could also be recovered from a method like
Annealed Importance Sampling (AIS) or Parallel Tempering (PT).

In a general form, some other distribution $q(\x) = \Ncal(\x ; \bmu, \bSigma)$ might also be available to us,
where the choice of $\bmu = \0$ and $\bSigma = \K$ would simply recover the prior.
For instance, if $q(\x)$ approximates $p(\x | \y)$
via some deterministic approximate inference method like Expectation Propagation (EP),
then this could be used as the starting distribution at $\beta = 0$ in AIS or PT.
We touch on AIS and PT in Section 6 of  \cite{fraccaro16adaptive}. 
For now,
let the unnormalised log density of $\x$ be\footnote{We include the normalising constants for $\prior(\x)$ and $q(\x)$ when
$\Lcal_{\beta}(\x)$ and its derivative are computed in Lines \ref{alg:derivatives:L1} and \ref{alg:derivatives:L2} in Algorithm \ref{alg:derivatives}, but as they're independent of $\x$,
we omit them here to make the explanation simpler.}
\begin{align}
\Lcal_{\beta}(\x) & = \beta \left[ \sum_{n=1}^{N} \ell_n(x_n) - \frac{1}{2} \x^T \K^{-1} \x \right] \nonumber \\
& \qquad\quad - (1 - \beta) \, \frac{1}{2}(\x - \bmu)^T \bSigma^{-1}(\x - \bmu) \label{eq:logjoint}
\end{align}
for any $\beta \in [0,1]$.
The generalised target distribution is
\begin{equation} \label{eq:pgeneral}
p_{\beta}(\x) \defined \frac{1}{Z(\beta)} \, \erm^{\Lcal_{\beta}(\x)}
\end{equation}
for any choice of $\beta$.
In this setup
$\beta = 1$ recovers $p(\x | \y)$, and $\beta = 0$ recovers $q(\x)$, or the prior if $q(\x) = \prior(\x)$.
Methods like AIS or PT estimate $\log Z$ by sampling from a sequence of distributions that range from $p_0$ to $p_1$.

In the rest of this note, after briefly introducing Hamiltonian Monte Carlo (HMC) in Section \ref{sec:hmc}, we will introduce RMHMC as an extension of HMC that exploits local second-order statistics (Section \ref{sec:rmhmc}). We will then discuss and evaluate a RMHMC method to simulate samples from (\ref{eq:pgeneral}).
Where the $\beta$ subscript is clear from the context, it will be dropped for brevity.


\section{Hamiltonian Monte Carlo}\label{sec:hmc}

\textit{Hamiltonian Monte Carlo (HMC)} \citep{Neal2010} can be used to define efficient proposal distributions for a Metropolis-Hastings sampler, that allow large moves in the parameter space while keeping a high acceptance rate.
It is particularly useful for eliminating the random walk behaviour that is typical of symmetric proposal distributions,
and improves poor mixing in case of highly correlated variables. 
The main idea behind this algorithm is to define an Hamiltonian function in terms of the target probability distribution, and move a sample from this distribution as if it was a particle in space following the corresponding Hamiltonian dynamics.

To sample a random variable $\x \in \mathbb{R}^N$ from the probability distribution $p(\x)$,
we introduce an independent auxiliary variable $\p \in \mathbb{R}^N$ with a Gaussian prior $p(\p)=\Ncal(\p;\mathbf{0},\M)$. Due to independence, the joint density can be written as
\begin{equation} \label{eq:hmc-joint}
p(\x,\p) = p(\x) \, p(\p) \propto \erm^{-H(\x, \p)} \ .
\end{equation}
The negative joint log-probability is then
\[ 
H(\x,\p) = -\Lcal(\x) + \frac{1}{2}\p^T\M^{-1}\p
\] 
and can be interpreted as a Hamiltonian $H(\x,\p)$ with potential energy $U(\x)$ and kinetic energy $K(\p)$,
\begin{align*}
U(\x) &= -\Lcal(\x) \\ 
K(\p) &= \frac{1}{2}\p^T\M^{-1}\p \ .
\end{align*}
The Hamiltonian represents the total energy of a closed system, in which $\x$ is the position of the particle in space.
Thanks to the quadratic kinetic term, the auxiliary variable $\p$ can be seen as a momentum variable, and the covariance matrix $\M$ as a mass matrix. 
As we will see in Section \ref{sec:rmhmc}, RMHMC improves the sampling efficiency of HMC by making use of a position-dependent mass matrix,
at the expense of complicating the overall algorithm.

To obtain samples from $p(\x)$, HMC simulates $\{ (\x^{(t)}, \p^{(t)}) \}_{t = 1}^{t_{\max}}$ samples from $p(\x, \p)$ and discards the $\p^{(t)}$ samples.
The remaining $\{ \x^{(t)} \}_{t = 1}^{t_{\max}}$ will then represent samples from the required marginal distribution.
HMC samples from the joint distribution $p(\x,\p)$ using a Gibbs sampling scheme with auxiliary variables $\p$.
\begin{enumerate}
\item Given the position and momentum $(\x^{(t)},\p^{(t)})$ at time $t$, the momentum is updated
by drawing a sample from the conditional distribution 
$p(\p^{(t+1)}|\x^{(t)})=p(\p^{(t+1)}) = \Ncal(\p^{(t+1)} ; \mathbf{0},\M)$.

\item Given $\p^{(t+1)}$,
the variable $\x^{(t+1)}$ is sampled from the conditional distribution $p(\x^{(t+1)} | \p^{(t+1)})$ with the Metropolis-Hastings algorithm, using a deterministic proposal distribution defined by simulating the behaviour of the physical system evolving under Hamiltonian dynamics and with initial position $(\x^{(t)},\p^{(t+1)})$. Given the Hamiltonian, we numerically integrate Hamilton's equations 
\begin{align}
\frac{\drm \x}{\drm \tau} & = \frac{\partial H}{\partial \p} = \frac{\partial K}{\partial \p}=\M^{-1}\p \nonumber \\
\frac{\drm \p}{\drm \tau} & = -\frac{\partial H}{\partial \x} = -\frac{\partial U}{\partial \x} =  \nabla_\x \Lcal(\x) \label{eq:hamiltonian-pde} \ .
\end{align}
and follow a trajectory to obtain a new pair ($\x^*$,$\p^*$), that is accepted with probability 
\begin{equation} \label{eq:acceptance}
\alpha = \min\left(1, \frac{\erm^{-H(\x^*,\p^*)}}{\erm^{-H(\x^{(t)},\p^{(t+1)})}} \right) \ .
\end{equation}
The accept-step appears in Lines \ref{alg:rmhmc:mcmc-if} to \ref{alg:rmhmc:mcmc-endif} in Algorithm \ref{alg:rmhmc}.
\end{enumerate}

The validity of this sampler relies on the \textit{reversibility} and \textit{volume preservation} properties of Hamiltonian mechanics \cite{Neal2010}.
When an analytic solution to the system of nonlinear differential equations is available, the proposed trajectory moves along the isocontours of the Hamiltonian in the phase space (and the acceptance rate in therefore one), whereas the random draws of the momentum $\p$ from the exact conditional distribution will change the energy levels.
For practical applications of interest,
Hamilton's equations do not have an analytic solution, and it is therefore necessary to discretise time and resort to numerical approximations. It is common to use the \textit{Stormer-Verlet leapfrog integrator}, that retains the reversibility and volume preservation properties required to obtain an exact sampler,
and computes the updates (in vector form) as
\begin{align}
\p(\tau + \tfrac{\epsilon}{2})& =\p(\tau) -\tfrac{\epsilon}{2}\nabla_\x U(\x(\tau)) \nonumber \\
\x(\tau + \epsilon)& =\x(\tau) + \epsilon \nabla_\p K(\p(\tau + \tfrac{\epsilon}{2})) \nonumber \\
\p(\tau + \epsilon) & = \p(\tau + \tfrac{\epsilon}{2}) -\tfrac{\epsilon}{2}\nabla_\x U(\x(\tau+\epsilon)) \ . \label{eq:hmc-leapfrog}
\end{align}
After half a step for the momentum variables, a full step for the position variables is done using the new momentum variables,
which is finally followed by the missing half step for the momentum variables using the updated position variables. This procedure is repeated $l_{\max}$ times for each sample to avoid random walk behaviour.
Due to the integration errors caused by the discretisation,
the Hamiltonian is not exactly conserved with the leapfrog method,
but it remains close to the true value to give a high acceptance probability. This error can be also controlled by careful tuning (manual or automatic) of the step size $\epsilon$ and the maximum number of integration steps $l_{\max}$.

\subsection{Tuning HMC}

The performance of Hamiltonian Monte Carlo is highly dependent on the correct tuning of the step size $\epsilon$, the number of leapfrog steps $l_{\max}$ done at each iteration and the mass matrix $\M$ of the momentum variable.
A too low step size $\epsilon$ makes it difficult to completely explore the whole space
unless a high $l_{\max}$ is used, but this causes an increase in computational time.
On the other hand, a too big $\epsilon$ can lead to an unstable algorithm that suffers from a low acceptance rate.
Furthermore, if $l_{\max}$ is too small there will be slow mixing due to random-walk behaviour, whereas a too high value for $l_{\max}$ may cause
\textit{double-back behaviour}, where the integrator returns to its starting point.
Picking the right mass matrix $\M$ is essential for optimised performance, as its diagonal terms have to reflect the scale of the sampled momentum variables and the off-diagonal terms their correlation: a simple default choice using a (possibly scaled) identity matrix will give poor results for highly correlated variables. 

It is finally worth noting that an acceptance rate of 1 is not the optimal choice, as this would mean for example that $\epsilon$ could be increased to move even further in the phase space. \cite{Neal2010} shows that the optimal acceptance rate should be around 0.65.
One can also consider a number of burn-in samples to avoid highly correlated proposed samples, not too high though for an efficient implementation.


\section{Riemannian Manifold Hamiltonian Monte Carlo}\label{sec:rmhmc}

\begin{figure*}[t]
\begin{center}
\includegraphics[width=0.7\textwidth]{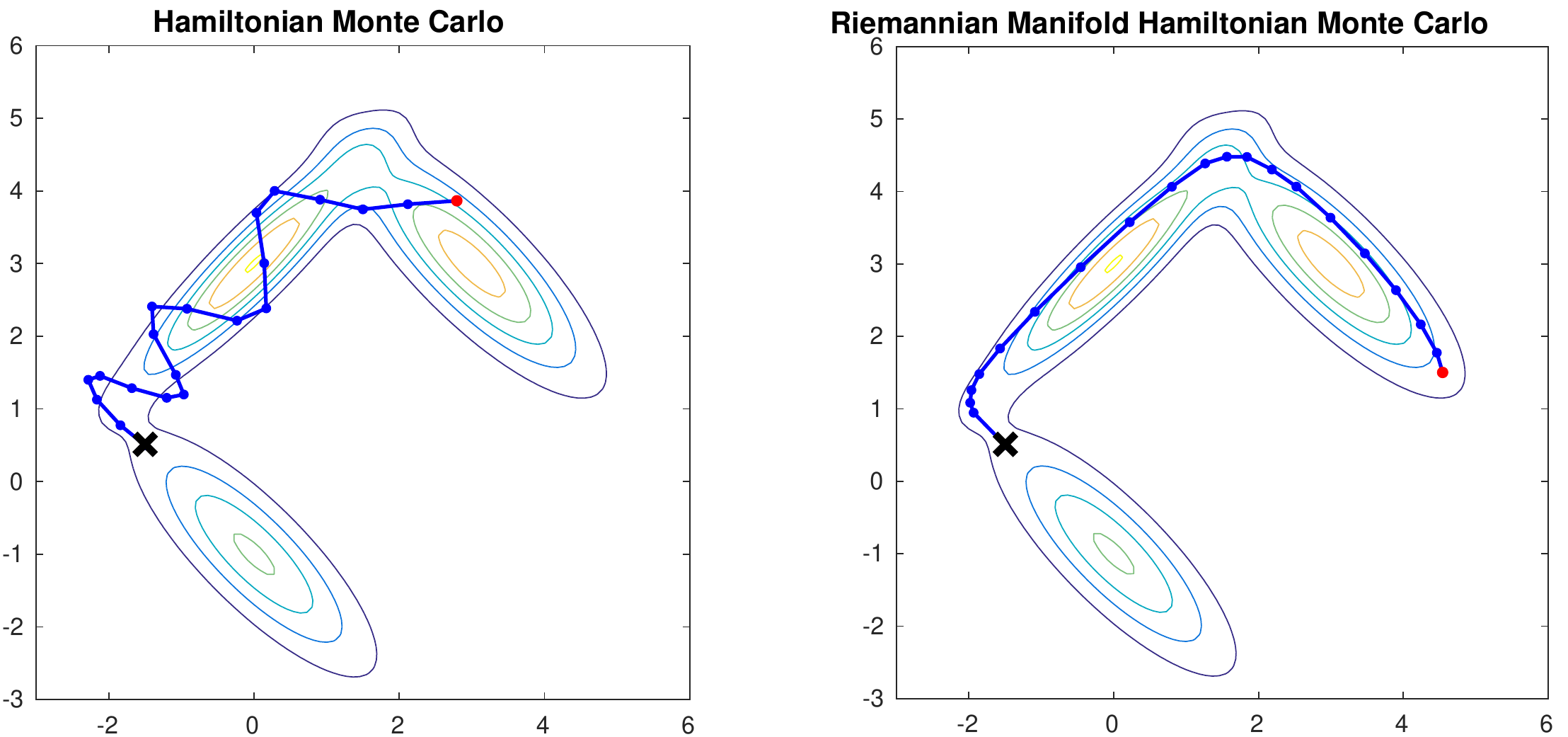}
\caption{Comparison between trajectories followed by HMC and RMHMC starting from the same initial position (the black cross).}
\label{fig:gmm}
\end{center}
\end{figure*}

As argued in the previous section, one of the main issues that arise when Hamiltonian Monte Carlo is used,
is the difficult tuning of the mass matrix $\M$, which is essential for good convergence of the sampler.
Girolami \emph{et al.}~show that
when, for instance, the position dependent expected Fisher information matrix $\G(\x)$ is used
instead of a fixed mass matrix $\M$,
many of the shortcomings of HMC can be addressed \cite{girolami09riemannian}.
The introduced algorithm -- \textit{Riemannian manifold Hamiltonian Monte Carlo (RMHMC)} -- can be seen as an extension of Hamiltonian Monte Carlo where the local geometry of the distribution we want to sample from is taken into account through the metric tensor $\G(\x)$.

In RMHMC, the covariance structure of the auxiliary Gaussian momentum variable is set as $\G(\x)$,
so that its distribution is shaped by the position of $\x$, i.e.~$p(\p | \x) = \Ncal(\p; \0, \G(\x))$.
While HMC factorises in (\ref{eq:hmc-joint}) and uses $p(\p) = \Ncal(\p ; \0, \M)$ for some fixed mass matrix $\M$, for RMHMC the joint density
\begin{equation} \label{eq:joint}
p(\x,\p) = p(\x) \, p(\p | \x) \propto \erm^{-H(\x, \p)}
\end{equation}
is no longer factorisable. As a result, the Hamiltonian
\begin{align}
H(\x,\p) & = -\Lcal(\x) + \frac{1}{2}\log\left(  (2\pi)^N \det(\G(\x)) \right) \nonumber \\
& \qquad + \frac{1}{2} \p^T \G(\x)^{-1}\p \label{eq:rmhmc-hamiltonian}
\end{align}
is not separable.
Unlike HMC, the term coming from the Gaussian normalising constant depends on $\x$, and it needs to be included
in the potential energy term.

RMHMC uses the same Gibbs sampler as HMC.
First, the momentum is sampled from the conditional distribution $\p|\x$ and then a new proposal for a Metropolis-Hastings sampler is found following a trajectory that is obtained by solving Hamilton's equations, that are in this case 
\begin{align}
\frac{\drm x_n}{\drm \tau} & = \frac{\partial H}{\partial p_n}= \{ \G(\x)^{-1} \p \}_n \nonumber \\
\frac{\drm p_n}{\drm \tau} & = -\frac{\partial H}{\partial x_n}=  \frac{\partial \Lcal(\x) }{ \partial x_n } - \frac{1}{2} \tr \left\{ \G(\x)^{-1} \frac{\partial \G(\x)}{ \partial x_n} \right \} \nonumber \\
& \qquad \quad\qquad+ \frac{1}{2} \p^T \G(\x)^{-1} \frac{\partial \G(\x) }{ \partial x_n } \G(\x)^{-1} \p \ .
\label{eq:rmhmc-hamiltonian}
\end{align}
Due to this dependence of the kinetic energy on the position (through $G(\x)$), the proposals generated from the leapfrog integrator will not satisfy detailed balance in a Hamiltonian Monte Carlo scheme.
To overcome this problem, \cite{girolami09riemannian} uses a more general leapfrog integrator, which is semi-explicit (i.e.~the update equations are defined implicitly and need to be solved with some fixed point iterations) but that satisfies reversibility and volume preservation,
therefore giving a correct sampler.
This \textit{generalised leapfrog integrator} leads to the following updates of the position and momentum variables:
\begin{align}
\p(\tau + \tfrac{\epsilon}{2}) & = \p(\tau) - \tfrac{\epsilon}{2} \nabla_\x H \Big(\x(\tau), \p(\tau+\tfrac{\epsilon}{2}) \Big)
\label{eq:integrator1} \\
\x(\tau + \epsilon) & = \x(\tau) + \tfrac{\epsilon}{2} \Bigg[ \nabla_\p H \Big(
\x(\tau), \p(\tau + \tfrac{\epsilon}{2}) \Big)
\label{eq:integrator2} \\
& \qquad \qquad  + \nabla_\p H \Big( \x(\tau + \epsilon), \p(\tau + \tfrac{\epsilon}{2})\Big) \Bigg] \nonumber \\
\p(\tau + \epsilon) & = \p(\tau + \tfrac{\epsilon}{2}) - \tfrac{\epsilon}{2} \nabla_\x H \Big(\x(\tau+\epsilon), \p(\tau + \tfrac{\epsilon}{2})\Big) \ . \label{eq:integrator}
\end{align}
It is simple to show that for separable Hamiltonians the generalised leapfrog integrator coincides with the one defined for HMC
in Equations (\ref{eq:hmc-leapfrog}).
When the Hamiltonian is non-separable, (\ref{eq:integrator1}) and (\ref{eq:integrator2}) are implicitly defined. In Algorithm
\ref{alg:rmhmc}, we solve them through \emph{fixed point iterations}:
Lines 13-17 for Equation (\ref{eq:integrator1}), and Lines 20-25 for Equation (\ref{eq:integrator2}).

Figure \ref{fig:gmm} shows an example of a trajectory followed by HMC and RMHMC to obtain one sample from a mixture of three Gaussians,
starting from the same initial position. We can see that RMHMC is better than HMC at taking into account the local geometry of the correlated Gaussian distributions, and can therefore do bigger moves in parameters space accepted with high probability.
 

\section{RMHMC for Gaussian Process Classification}

\begin{algorithm*}[!t]
\begin{algorithmic}[1]
\STATE \textbf{input:}
$\y$, $\K$, $\L^{\mathrm{K}}$, $\log |\K|$, $\bmu$, $\widetilde{\bSigma}$,
$\L^{\Sigma}$, $\log |\bSigma|$, $\beta$, $\x_0$, $l_{\max}$, $\epsilon$, $t_{\max}$
\STATE $\x := \x_0$
\STATE $\{ \Lcal, \frac{\partial \Lcal}{\partial \x}, \G^{-1}, \log |\G|, \drm \g \} := \algofunc{riemann-metric}
(\x, \y, \K, \L^{\mathrm{K}}, \log |\K|, \bmu, \widetilde{\bSigma}, \L^{\Sigma}, \log |\bSigma|, \beta)$  
\FOR {$t=1$ to $t_{\max}$}
	\STATE $\L^{\mathrm{Ginv}} = \chol(\G^{-1})$            
	\STATE $\p \sim \Ncal(\p ; \0, \I)$
	\STATE $\p := (\L^{\mathrm{Ginv}})^T \backslash \p$	\hfill \greentext{\emph{// initial momentum is} $\p \sim \Ncal(\p ; \0, \G(\x)^{-1})$}
	\STATE $\{ H, \frac{\partial H}{\partial \x} \} :=
			\algofunc{hamiltonian-and-gradient}
	        (\Lcal, \frac{\partial \Lcal}{\partial \x}, \G^{-1}, \log |\G|, \drm \g, \p)$    
	\STATE $H^{\old} := H; \,
			\x^{\old} := \x; \,
			{\frac{\partial \Lcal}{\partial \x}}^{\old} := \frac{\partial \Lcal}{\partial \x}; \,
			\Lcal^{\old} := \Lcal; \,
			(\G^{-1})^{\old} := \G^{-1}; \,
			(\log |\G|)^{\old} := \log |\G|; \,
			(\drm \g)^{\old} := \drm \g$
	\STATE \greentext{\emph{// take $l_{\max}$ leapfrog steps:}}
	\FOR {$l = 1$ to $l_{\max}$}
		\STATE \greentext{\emph{// the fixed point loops $f = 1, \ldots, f_{\max}$ below are a time-reversible volume preserving numerical integrator for solving the non-separable Hamiltonian to ensure a correct MCMC algorithm}}
		\FOR {$f = 1$ to $f_{\max}$}
			\STATE $\p' := \p - \frac{1}{2} \epsilon \frac{\partial H}{\partial \x}$ \hfill \greentext{{\it // equation (\ref{eq:integrator1})}}
			\STATE $\{ H, \frac{\partial H}{\partial \x} \} :=
			        \algofunc{hamiltonian-and-gradient}
			        (\Lcal, \frac{\partial \Lcal}{\partial \x}, \G^{-1}, \log |\G|, \drm \g, \p')$
		\ENDFOR
		\STATE $\p := \p'$
       \STATE $\frac{\partial H}{\partial \p} := \G^{-1} \p$
		\STATE $\frac{\partial H}{\partial \p}' := \frac{\partial H}{\partial \p}$
		\FOR {$f = 1$ to $f_{\max}$}
			\STATE $\x' := \x + \frac{1}{2} \epsilon (\frac{\partial H}{\partial \p} + \frac{\partial H}{\partial \p}')$ \hfill \greentext{{\it // equation (\ref{eq:integrator2})}}  			
			\STATE $\{ \Lcal, \frac{\partial \Lcal}{\partial \x}, \G^{-1}, \log |\G|, \drm \g \} :=			
                    \algofunc{riemann-metric}
                    (\x', \y, \K, \L^{\mathrm{K}}, \log |\K|, \bmu, \widetilde{\bSigma}, \L^{\Sigma}, \log |\bSigma|, \beta)$  
			\STATE $\frac{\partial H}{\partial \p}' := \G^{-1} \p$                                   
		\ENDFOR
		\STATE $\x := \x'$
		\STATE $\{ H, \frac{\partial H}{\partial \x} \} :=
		        \algofunc{hamiltonian-and-gradient}
		        (\Lcal, \frac{\partial \Lcal}{\partial \x}, \G^{-1}, \log |\G|, \drm \g, \p)$
		\STATE $\p := \p - \frac{1}{2} \epsilon \frac{\partial H}{\partial \x}$ \hfill \greentext{{\it // equation (\ref{eq:integrator})}}           
		\STATE $\{ H, \frac{\partial H}{\partial \x} \} :=
		        \algofunc{hamiltonian-and-gradient}
		        (\Lcal, \frac{\partial \Lcal}{\partial \x}, \G^{-1}, \log |\G|, \drm \g, \p)$
	\ENDFOR
	\IF  {$\mathrm{rand} < \exp\{H^{\old} - H \}$} \label{alg:rmhmc:mcmc-if}
		\STATE \greentext{\emph{// accept; do nothing}}
	\ELSE
		\STATE \greentext{\emph{// reject}}
	\STATE $\x := \x^{\old}; \,
			\frac{\partial \Lcal}{\partial \x} := {\frac{\partial \Lcal}{\partial \x}}^{\old}; \,
			\Lcal := \Lcal^{\old}; \,
			\G^{-1} := (\G^{-1})^{\old}; \,
			\log |\G| := (\log |\G|)^{\old}; \,
			\drm \g := (\drm \g)^{\old}$
	\ENDIF \label{alg:rmhmc:mcmc-endif}
	\STATE $\mathrm{samples}(t) = \x$
	\STATE $\mathrm{energies}(t) = - \Lcal$
\ENDFOR 
\STATE \textbf{return} samples, energies
\end{algorithmic}
\caption{Riemannian Manifold Hamiltonian Monte Carlo}
\label{alg:rmhmc}
\end{algorithm*}


In this section, we take GP classification (GPC) as a working example of a GP model with non-Gaussian likelihood terms.
For a GPC problem, every latent function value $x_n$ supports a binary observation $y_n \in \{ -1, +1 \}$.
The probit function $\Phi(x) = \int_{-\infty}^x \Ncal(z ; 0, 1) \, \drm z$ is often used to
model the likelihood,
\[
\ell_n(x_n) \defined \log p(y_n | x_n) = \log \Phi(y_n x_n) \ .
\]
The resulting function $\Lcal_{\beta}(\x)$ in (\ref{eq:logjoint}) is concave, and $p(\x | \y)$ is usually very correlated.

The Markov chain Monte Carlo (MCMC) algorithm that draws samples from
(\ref{eq:pgeneral}) via the joint density in (\ref{eq:joint}) is given in pseudocode in Algorithm~\ref{alg:rmhmc}.
We'll first discuss the its inputs in Section \ref{sec:inputs}, and then turn to its choice of metric tensor
in Section \ref{sec:metrictensor}.
The resulting Hamiltonian and its gradients are given in Section \ref{sec:hamiltonian-and-gradients}.
Section \ref{sec:log-likelihood-derivatives} gives the first, second, and third derivatives for a probit log
likelihood, which are all required in $\partial H / \partial \x$.

\subsection{Inputs} \label{sec:inputs}

Algorithm~\ref{alg:rmhmc} requires the parameters of $\Lcal_{\beta}(\x)$ in (\ref{eq:logjoint}), with minimal other external settings:
\begin{itemize}
\item The input observations $\y$ from (\ref{eq:loglikelihood}) and kernel matrix $\K$ from the GP prior in (\ref{eq:gpprior}) are required.
It is also useful to precompute the Cholesky decomposition $\K = \L^{\mathrm{K}} (\L^{\mathrm{K}})^T$ and log determinant $\log |\K|$ of the kernel matrix.
\item If $q(\x) = \Ncal(\x ; \bmu, \bSigma)$ is an approximation obtained by EP,
then the precision matrix of $q$ decomposes as $\K^{-1}$
plus a \emph{diagonal} matrix $\widetilde{\bSigma}$ containing the contributions from $N$ approximate 
factors corresponding to the likelihood terms,
\[
\bSigma^{-1} = \K^{-1} + \widetilde{\bSigma}^{-1} \ .
\]
We choose the above notation to match that of the GP classification approximation in Rasmussen and Williams's
\emph{Gaussian Processes for Machine Learning} book \cite{rasmussen06book}.
Of course $\bSigma$ needn't have this form, but as we consider an EP approximation in Section \ref{sec:results},
use this form for numerical stability.
In addition to taking $\bmu$ and $\widetilde{\bSigma}$ as inputs, we also
precompute the Cholesky decomposition
$\bSigma = \L^{\Sigma} (\L^{\Sigma})^T$ and its log determinant $\log |\bSigma|$.
\item Inverse temperature $\beta$, a starting point $\x_0$, 
the number of leapfrog steps $l_{\max}$ per sample,
a step-size $\epsilon$, and
a maximum number of samples $t_{\max}$ are also required.
We implicitly assume that a burn-in sample would be discarded from $\{ \x_t \}_{t = 1}^{t_{\max}}$,
and that the number of fixed point iterations is pre-set to, say, $f_{\max} = 5$.
\end{itemize}


\begin{algorithm*}[t]
\begin{algorithmic}[0]
\STATE \textbf{function} \algofunc{hamiltonian-and-gradient}
\STATE \textbf{input:}
$\Lcal$,
$\, \frac{\partial \Lcal}{\partial \x}$, $\, \G^{-1}$,
$\, \log |\G|$,
$\, \drm \g$,
$\, \p$ \hfill \greentext{\it $\drm \g \defined \mathsf{vec} [ \frac{\partial \G(\x)_{nn}}{\partial x_n} ]$}
\STATE $H = \frac{1}{2} \p^T \G^{-1} \p + \frac{1}{2} \log |\G| - \Lcal$ \hfill \greentext{\it // equation (\ref{eq:rmhmc-hamiltonian}), but ignoring $\tfrac{N}{2}\log (2 \pi)$ as a cancelling constant in (\ref{eq:acceptance})}
\STATE $\w = \G^{-1} \p$    
\STATE $\frac{\partial H}{\partial \x} = 
\frac{1}{2} (\drm \g \circ \mathsf{diag}(\G^{-1})) - \frac{1}{2}(\w \circ \w \circ \drm \g) - \frac{\partial \Lcal}{\partial \x}$    
\STATE \textbf{return} $\{ H, \frac{\partial H}{\partial \x} \}$
\end{algorithmic}
\caption{Riemann Hamiltonian and Gradient}
\label{alg:riemannhamiltonian}
\end{algorithm*}


\begin{algorithm*}[tb]
\begin{algorithmic}[1]
\STATE \textbf{function} \algofunc{riemann-metric}
\STATE \textbf{input:}
$\x$, $\, \y$, $\, \K$, $\, \L^{\mathrm{K}}$, $\, \log |\K|$,
$\, \bmu$, $\, \widetilde{\bSigma}$, $\, \L^{\Sigma}$, $\, \log |\bSigma|$, $\, \beta$
\STATE $\{ \Lcal, \frac{\partial \Lcal}{\partial \x} \} :=
        \algofunc{derivatives}
        (\x, \y, \L^{\mathrm{K}}, \log |\K|, \bmu, \L^{\Sigma}, \log |\bSigma|, \beta)$
\STATE \greentext{\it // compute the non-zero (diagonal) entries of $\frac{\partial \G(\x)}{\partial x_n}$ as a vector $\drm \g$:}
\STATE $\z := \y \circ \x$
\STATE $\r := - \frac{1}{2} \z^2 - \frac{1}{2} \log(2 \pi) - \log \Phi(\z)$
       \hfill \greentext{\it // $\r$ is the log ratio of $\Ncal(yx) / \Phi(yx)$, and $\z^2 \defined \z \circ \z$, i.e.~element-wise square}
\STATE $\blambda := \z \circ \exp\{ \r \} + \exp\{ 2 \r\}$
       \hfill \greentext{\it// $\exp\{ \r \} \defined \vec[\exp\{ r_n\}]$, i.e.~applied element-wise}
\STATE $\blambda := \beta \blambda$     \label{alg:riemannmetric:lambda}
\STATE $\drm \g := \y \circ (1 - \x^2) \circ \exp\{\r\} - 3 \x \circ \exp\{2 \r\} - 2 \y \circ \exp\{3 \r\}$ 
\STATE $\drm \g := \beta (\drm \g)$     \label{alg:riemannmetric:dg}
\STATE \greentext{\it // compute the inverse $\G(\x)^{-1}$ and its determinant:}
\STATE $\s := \sqrt{\blambda}$
       \hfill \greentext{\it// element-wise root}
\IF{$\bSigma = \K$ (i.e.~$\widetilde{\bSigma} = \0$) and $\bmu = \0$} \label{alg:riemannmetric:if}
	\STATE $\L := \chol(\I + \K \circ (\s \s^T))$
           \hfill \greentext{\it // slower is $\S := \diag(\s)$ and then $\L := \chol(\I + \S \K \S)$}
	\STATE $\V = \L \backslash (\K \star \s)$
           \hfill \greentext{\it // slower is $\V = \L \backslash (\S \K)$}
    \STATE \greentext{\it // $\star$ indicates the row-wise product between $\K$ and $\s$, multiplying $s_1$ with the first row of $\K$, $s_2$ with $\K$'s second row, etc.}
	\STATE $\G^{-1} := \K - \V^T \V$;
	\STATE $\log |\G| := - 2 \sum_{n=1}^{N} \log L^{\mathrm{K}}_{nn} + 2 \sum_{n=1}^{N} \log L_{nn}$
\ELSE  \label{alg:riemannmetric:else} 
	\STATE $\t := \sqrt{1 - \beta} \, \diag(\widetilde{\bSigma}^{-\frac{1}{2}})$
	\STATE $\L := \chol(\I + \K \circ (\t \t^T))$
		   \hfill \greentext{\it // slower is $\T := \diag(\t)$ and then $\L := \chol(\I + \T \K \T)$}
		   \label{alg:riemannmetric:LIKTT}
	\STATE $\V := \L \backslash (\K \star \t)$
		   \hfill \greentext{\it // slower is $\V = \L \backslash (\T \K)$}
	\STATE $\A = \K - \V^T \V$
           \hfill \greentext{\emph{// using $\A^{-1} = \K^{-1} + (1 - \beta) \widetilde{\bSigma}^{-1}$}}
	\STATE $\log |\A^{-1}| = - 2 * \sum_{n=1}^{N} \log L_{nn}^{\mathrm{K}} + 2 \sum_{n=1}^{N} \log L_{nn}$
	\STATE $\L := \chol(\I + \A \circ (\s \s^T))$
	       \hfill \greentext{\it // slower is $\S := \diag(\s)$ and then $\L := \chol(\I + \S \A \S)$}
	\STATE $\V := \L \backslash (\A \star \s)$
	       \hfill \greentext{\it // slower is $\V = \L \backslash (\S \A)$}
	\STATE $\G^{-1} := \A - \V^T \V$
           \hfill \greentext{\emph{// using $\G = \bLambda + \A^{-1}$}}
	\STATE $\log |\G| := \log |\A^{-1}| + 2  \sum_{n=1}^{N} \log L_{nn}$
\ENDIF
\STATE \textbf{return} $\{ \Lcal, \frac{\partial \Lcal}{\partial \x}, \G^{-1}, \log |\G|, \drm \g \}$
\end{algorithmic}
\caption{Metric Tensor}
\label{alg:riemannmetric}
\end{algorithm*}


\begin{algorithm*}[tb]
\begin{algorithmic}[1]
\STATE \textbf{function} \algofunc{derivatives}
\STATE \textbf{input:}
$\x$, $\, \y$, $\, \L^{\mathrm{K}}$, $\, \log |\K|$,
$\, \bmu$, $\, \L^{\Sigma}$, $\, \log |\bSigma|$, $\, \beta$
\STATE \greentext{\emph{// compute the value of the log likelihood for the probit classification model and its derivative with respect to $\x$}}
\STATE $\z := \y \circ \x$
\STATE $L := \sum_{n=1}^{N} \log \Phi(z_n)$
\STATE $\Ical := \mathsf{indexes}( z_n > -15 )$
       \hfill \greentext{\emph{// normal regime; $\neg \Ical$ indexes asymptotic regime for numeric stability}}
\STATE $\frac{\partial L}{\partial \x}(\Ical) := (2 \pi)^{-1/2} \exp \{- \frac{1}{2} \z_{\Ical}^2 \} / \Phi(\z_{\Ical})$
       \hfill \greentext{\emph{// normal regime; element-wise division}}
\STATE $\frac{\partial L}{\partial \x}(\neg \Ical) := - \z_{\neg \Ical} - 1 / \z_{\neg \Ical} + 2 / \z_{\neg \Ical}^3$
       \hfill \greentext{\emph{// asymptotic regime; element-wise division}}
\STATE $\frac{\partial L}{\partial \x} := \y \circ \frac{\partial L}{\partial \x}$ \label{alg:derivatives:Lx}
\STATE \greentext{\emph{// add the derivatives with respect to the log prior}}
\STATE $\f = \L^{\mathrm{K}} \backslash \x$
\IF{$\L^{\Sigma} = \L^{\mathrm{K}}$ and $\bmu = \0$}
	\STATE $\Lcal := \beta L - \frac{N}{2} \log(2 \pi) - \frac{1}{2} \log |\K| - \frac{1}{2} \f^T \f$ \hfill \greentext{\it // equation (\ref{eq:logjoint})} \label{alg:derivatives:L1}
	\STATE $\frac{\partial \Lcal}{\partial \x} := \beta \frac{\partial L}{\partial \x} - ((\L^{\mathrm{K}})^T \backslash \f)$
\ELSE
	\STATE $\f' := \L^{\Sigma} \backslash (\x - \bmu)$
	\STATE $\Lcal := \beta L
- \frac{N}{2} \log(2 \pi)
- \beta( \frac{1}{2} \log |\K| + \frac{1}{2} \f^T \f)
- (1 - \beta)( \frac{1}{2} \log |\bSigma| + \frac{1}{2} {\f'}^T \f')$ \hfill \greentext{\it // equation (\ref{eq:logjoint})} \label{alg:derivatives:L2}
	\STATE $\frac{\partial \Lcal}{\partial \x} :=
  \beta \frac{\partial L}{\partial \x} + \beta ((\L^{\mathrm{K}})^T \backslash \f)
- (1 - \beta) ((\L^{\Sigma})^T \backslash \f')$
\ENDIF
\STATE \textbf{return} $\{ \Lcal, \frac{\partial \Lcal}{\partial \x} \}$
\end{algorithmic}
\caption{Derivatives of $\Lcal_{\beta}(\x)$ in Equation (\ref{eq:logjoint})}
\label{alg:derivatives}
\end{algorithm*}

\subsection{Metric tensor} \label{sec:metrictensor}

There are many choices of a metric for a specific manifold, and a more detailed discussion beyond the scope of this note
is presented by Girolami and Calderhead \cite{RMHMC}. We choose the negative second derivative
of $\Lcal(\x)$ in (\ref{eq:logjoint}), evaluated at $\x$, as metric tensor:
\[
\G(\x) \defined - \nabla_{\x}^2 \, \Lcal(\x) \ .
\]
Therefore $\G(\x)$ is simply a Hessian matrix,
\[
- \nabla_{\x}^2 \, \Lcal(\x) 
= \underbrace{ - \beta \, \mathsf{diag} \Big[ \nabla_{x_n}^2 \ell_n(x_n) \Big] }_{\bLambda(\x)} + \beta \K^{-1} 
+ (1 - \beta) \bSigma^{-1} .
\]
Notation $\diag$ indicates a diagonal matrix formed by its arguments.
We deliberately set aside a definition of $\bLambda(\x)$ as the only component of $\G(\x)$
that depends on $\x$. This simplifies later derivations, for which another derivative (the third)
is required in $\partial \G(\x) / \partial x_n$ in (\ref{eq:rmhmc-hamiltonian}) when simulating the Hamiltonian dynamics.
Hence
\begin{equation}
\G(\x) = \bLambda(\x) + \beta \K^{-1} + (1 - \beta) \bSigma^{-1} .
\end{equation}

\subsection{The Hamiltonian and its gradients}
\label{sec:hamiltonian-and-gradients}

First consider $- \partial H / \partial \x$ in Equation (\ref{eq:rmhmc-hamiltonian}).
A fast way to compute it
is given in function \algofunc{hamiltonian-and-gradient} in Algorithm~\ref{alg:riemannhamiltonian}.
It relies on the metric tensor, computed by the \algofunc{riemann-metric} function in Algorithm~\ref{alg:riemannmetric}, and the derivatives $\nabla_{\x} \Lcal(\x)$,
computed in the \algofunc{derivatives} function in Algorithm~\ref{alg:derivatives}.
Considering the latter, the derivative of $\Lcal$ with respect to $\x$ is
\begin{align*}
\nabla_{\x} \Lcal(\x)
& = \beta \, \vec \Big[ \nabla_{x_n} \ell_n(x_n) \Big] - \beta \K^{-1} \x \\
& \qquad\qquad - (1 - \beta) \bSigma^{-1} (\x - \bmu)
\end{align*}
where $\vec$ gives a column vector of its arguments.
Function \algofunc{derivatives} uses pre-computed Cholesky decompositions of
$\K = \L^{\mathrm{K}} (\L^{\mathrm{K}})^T$ and $\bSigma = \L^{\Sigma} (\L^{\Sigma})^T$
to stably determine $\K^{-1} \x$ and $\bSigma^{-1} (\x - \bmu)$ through back-solving.
The Cholesky factors are lower-diagonal, and $\circ$ indicates the element-wise
or Hadamard product between a pair of vectors or matrices.

Turning to the \algofunc{riemann-metric} function, quantities like 
\[
\tr \left\{ \G(\x)^{-1} \frac{\partial \G(\x)}{\partial x_n} \right\}
\]
are required in Equation (\ref{eq:rmhmc-hamiltonian}), whilst the evaluation of $H$
also needs $\log |\G(\x)|$.
Notice that the derivative
\begin{equation} \label{eq:G-der}
\left\{ \frac{\partial \G}{\partial x_n} \right\}_{n,n} = \frac{\partial \Lambda_{nn}}{\partial x_n}
\end{equation}
is non-zero in position $(n,n)$, and zero elsewhere, and $\G$ has no other dependence on $\x$ than through the log likelihood derivatives. We therefore only keep the non-zero entries, given by (\ref{eq:G-der}), in a vector $\drm \g$ in 
Algorithm~\ref{alg:riemannmetric}.

To evaluate the log determinant $\log |\G(\x)|$ and the inverse $\G(\x)^{-1}$,
we'll consider the cases where $q(\x) = p(\x)$ and where $q(\x) \neq p(\x)$ separately.
Looking at \algofunc{riemann-metric}, their evaluations
are under the two branches, starting from Lines \ref{alg:riemannmetric:if} and \ref{alg:riemannmetric:else},
of its only if-then-else fork.
To speed up computation,
we introduce an additional operator: Let $\star$ indicate a row-wise product
between a matrix and a vector, e.g.~$\K \star \s$ multiplies $s_1$ with the first row of $\K$, $s_2$ with the second row, etc.
If $\S \defined \diag(\s)$, the result is equal to (but faster to compute than) the matrix product $\S \K$.

\subsubsection{The inverse $\G(\x)^{-1}$ and determinant $\log |\G(\x)|$ when $q(\x) = p(\x)$}

If $q$ is equal to the prior, then
\[
\G(\x)^{-1} = (\bLambda(\x) + \K^{-1})^{-1} \ ,
\]
for which we'll use the Sherman-Morrison-Woodbury formula. We'll expand the steps in more detail
in Section \ref{eq:q-not-p} below. Briefly, the steps are
to let $\s := \diag(\bLambda^{\frac{1}{2}})$, to determine a stable
Cholesky decomposition in $\L^{\mathrm{s}} = \chol(\I + \K \circ (\s \s^T))$
(note that $\K \circ (\s \s^T) = \S \K \S$, but that the former is faster to compute),
and to use $\L^{\mathrm{s}}$ to back-solve $\V = \L^{\mathrm{s}} \backslash (\K \star \s)$.
Then\footnote{ As in the \algofunc{riemann-metric} function in Algorithm~\ref{alg:riemannmetric}, we use $\V$ as a ``local variable'' in this section, with a similar ``local'' use in Section \ref{eq:q-not-p}. }
\[
\G(\x)^{-1} := \K - \V^T \V \ ,
\]
and using the precomputed Cholesky decomposition $\L^{\mathrm{K}} = \chol(\K)$ of the kernel matrix,
\[
\log |\G(\x)| := - 2 \sum_{n=1}^{N} \log L^{\mathrm{K}}_{nn} + 2 \sum_{n=1}^{N} \log L_{nn}^{\mathrm{s}} \ .
\]

\subsubsection{The inverse $\G(\x)^{-1}$ and determinant $\log |\G(\x)|$ when $q(\x) \neq p(\x)$} \label{eq:q-not-p}

To determine the determinant when $q$ is not equal to the prior, we define
\[
\A^{-1} \defined \beta \K^{-1} + (1 - \beta) \bSigma^{-1}
\]
so that $\G(\x)^{-1} = (\bLambda(\x) + \A^{-1})^{-1}$.
If $q$ is given by an EP approximation, then the precision matrix of $q$ decomposes as $\K^{-1}$
plus a \emph{diagonal} matrix $\widetilde{\bSigma}$ containing the contributions from $N$ approximate 
factors corresponding to the likelihood terms,
\[
\bSigma^{-1} = \K^{-1} + \widetilde{\bSigma}^{-1} \ .
\]
We choose the above notation to match that of the GPC approximation in Rasmussen and Williams's
\emph{Gaussian Processes for Machine Learning} book.
Of course $\bSigma$ needn't have this form, but as we use an EP approximation,
we utilize it in aid of numerical stability.
Then
\[
\A^{-1} \defined \K^{-1} + (1 - \beta) \widetilde{\bSigma}^{-1} = \K^{-1} + \B
\]
for diagonal positive definite matrix $\B = (1 - \beta) \widetilde{\bSigma}^{-1}$.
In Line \ref{alg:riemannmetric:LIKTT} in Algorithm \ref{alg:riemannmetric} we use
\[
\T \defined \B^{ \frac{1}{2}} \ ,
\]
and will do so below. To determine the inverse of $\A$, the Woodbury identity states
\begin{align*}
\A & = (\K^{-1} + \T \T)^{-1} \\
& = \K - \K \T
( \underbrace{ \I + \T \K \T }_{\L^{\mathrm{t}} (\L^{\mathrm{t}})^T} )^{-1} \T \K
\ ,
\end{align*}
with a Cholesky decomposition $\L^{\mathrm{t}} \defined \chol(\I + \T \K \T)$.
The steps are to set
$\t := \sqrt{1 - \beta} \, \diag( \widetilde{\bSigma}^{-\frac{1}{2}})$
as the vector, and then to determine
$\L^{\mathrm{t}} := \chol(\I + \K \circ (\t \t^T))$,
$\V := \L^{\mathrm{t}} \backslash (\K \star \t)$, and $\A = \K - \V^T \V$.
The log determinant of $\A$'s inverse follows from a similar identity,
\begin{align*}
\log |\A^{-1}| & = \log |\K^{-1} + \T \T| \\
& = \log |\I| + \log |\K^{-1}| + \log | \I + \T \K \T | \ ,
\end{align*}
which will re-use the Cholesky decomposition of $\I + \T \K \T$
to give
\[
\log |\A^{-1}| = - 2 \sum_{n=1}^{N} \log L_{nn}^{\mathrm{K}} + 2 \sum_{n=1}^{N} \log L_{nn}^{\mathrm{t}} \ .
\]
The same set of steps can be repeated to find $\G(\x)^{-1}$ and its determinant.
As $\G(\x) = \bLambda(\x) + \A^{-1}$,
let $\L^{\mathrm{a}} := \chol(\I + \A \circ (\s \s^T))$, with $\s$ defined as before.
With $\V := \L^{\mathrm{a}} \backslash (\A \star \s)$,
we obtain $\G(\x)^{-1} := \A - \V^T \V$
and
\[
\log |\G(\x)|: = \log |\A^{-1}| + 2  \sum_{n=1}^{N} \log L_{nn}^{\mathrm{a}} \ .
\]

\subsection{Derivatives of a probit log likelihood}
\label{sec:log-likelihood-derivatives}

The first, second, and third derivatives of the log likelihood are required in RMHMC.
All derivatives appears in the gradient of $H(\x, \p)$ with respect to $\x$ in Equation (\ref{eq:rmhmc-hamiltonian}):
the first derivate in the gradient of $\Lcal(\x)$,
the second derivative in the metric tensor $\G(\x) \defined - \nabla_{\x}^2 \Lcal(\x)$,
and the third derivative in the gradient of the metric tensor $\partial \G(\x) / \partial x_n$.

The log likelihood for $x_n$, on observing $y_n \in \{ -1, +1 \}$, is the log probit function
\[
\ell (x) = \log p(y | x) = \log \Phi(y x) \ ,
\]
where subscripts $n$ are dropped for brevity.
Let $\Ncal(y x) \defined \Ncal(yx ; 0, 1)$ denote a centred unit-variance Gaussian. Then the first derivative
\[
\nabla_{x} \ell(x) = \frac{\Ncal(y x)}{\Phi(y x)}
\]
is given in Line \ref{alg:derivatives:Lx} in the \algofunc{derivatives} function in Algorithm \ref{alg:derivatives}.

Taking the derivative again (and multiplying by $-1$), the diagonal values of $\bLambda(\x)$ are
\[
\Lambda_{nn} = \beta \left[ y x \, \frac{\Ncal(y x)}{\Phi(y x)} + \left( \frac{\Ncal(y x)}{\Phi(y x)}  \right)^2 \right] \ ,
\]
and are computed as a vector $\blambda$ in Line \ref{alg:riemannmetric:lambda} in the \algofunc{riemann-metric} function in Algorithm
\ref{alg:riemannmetric}.

In the derivative of the metric tensor $\partial \G(\x) / \partial x_n$, the log likelihood contributes
\begin{align*}
\frac{ \partial \Lambda_{nn} }{ \partial x_n}
& = \beta \left[ y (1 - x^2) \, \frac{\Ncal(y x)}{\Phi(y x)}
- 3 x \left( \frac{\Ncal(y x)}{\Phi(y x)}\right)^2 \right. \\
& \qquad\qquad \left. - 2 y \left( \frac{\Ncal(y x)}{\Phi(y x)}\right)^3 \right] ,
\end{align*}
and is computed as a vector $\drm \g$ in Line \ref{alg:riemannmetric:dg} in the \algofunc{riemann-metric} function.


\section{Results} \label{sec:results}

\begin{figure*}[t!]
\begin{center}
\includegraphics[width=0.9\textwidth]{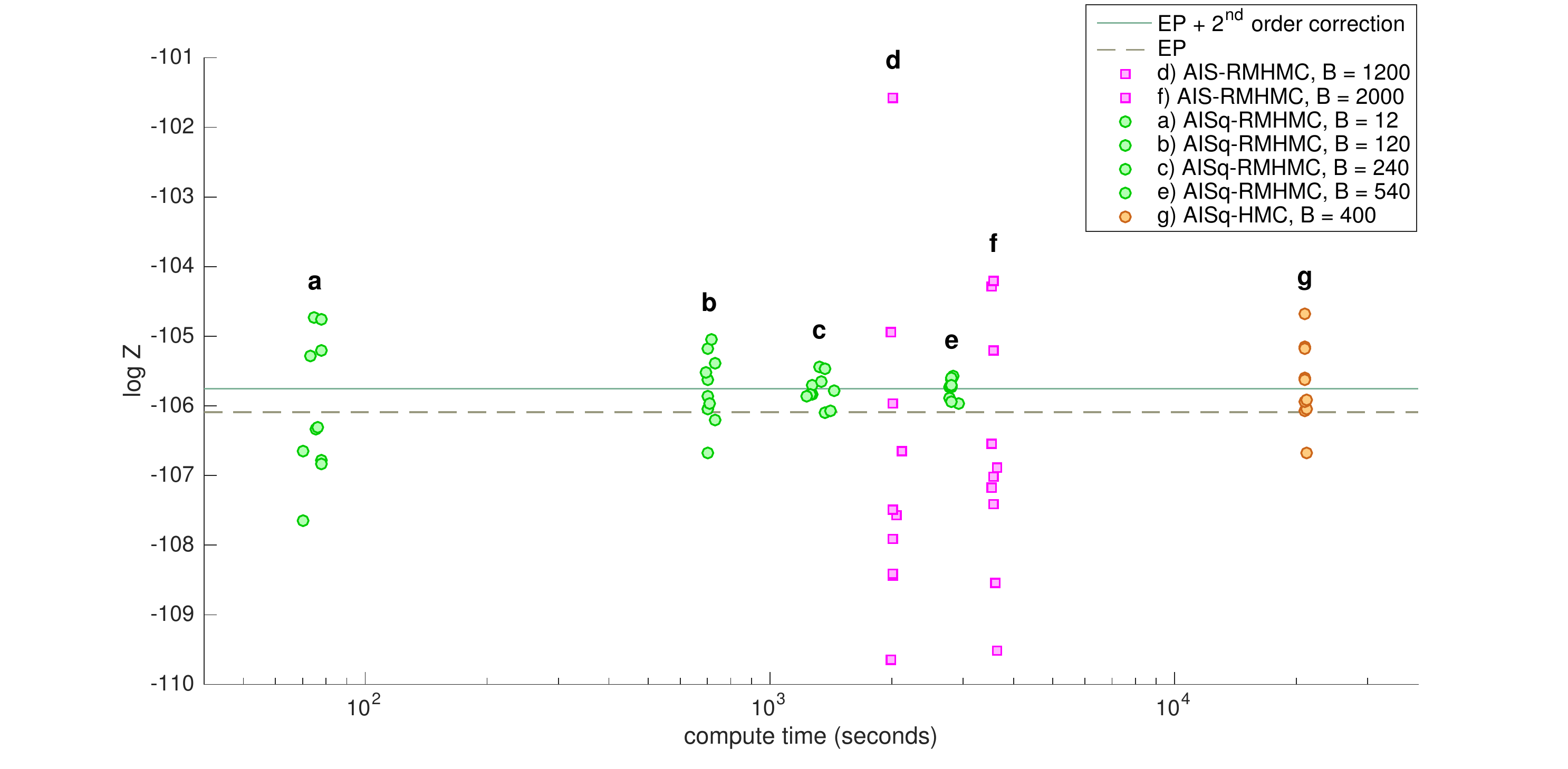}
\caption{
An extensive time-based comparison for GPC on the USPS 3-vs.-5 data set, using a highly correlated prior. From left to right, the evaluations are for
(a) AIS$q$-RMHMC using $B=12$;
(b) AIS$q$-RMHMC using $B = 120$;
(c) AIS$q$-RMHMC using $B = 240$;
(d) AIS-RMHMC using $B = 1200$;
(e) AIS$q$-RMHMC using $B = 540$;
(f) AIS-RMHMC using $B = 2000$;
(g) AIS$q$-HMC using $B = 400$.
The dotted line indicates EP's approximation, and the solid line a second order correction to the EP solution.
A label that is starred indicates that an AIS method was aided by annealing from EP's $q(\x)$ to $p(\x | \y)$, and not from the prior, as is more commonly done.
}
\label{fig:gpc}
\end{center}
\end{figure*} 

The results presented here are form a baseline for the \emph{Adaptive Resample-Move} algorithm in \cite{fraccaro16adaptive}.
The evaluation is on the  USPS 3-vs.-5 data set \cite{KussAssessing},
using a covariance function $K_{mn} = k(\bxi_m, \bxi_n) = \sigma^2 \exp(-\frac{1}{2} \| \bxi_m - \bxi_n\|^2 / \ell^2 )$ that correlates inputs $\bxi_m$ and $\bxi_n$ through a length scale $\ell = \exp(4.85)$ and amplitude parameter $\sigma = \exp(5.1)$.\footnote{On \cite{KussAssessing}'s entire $(\log \ell, \log \sigma)$-grid, this setting proved to be the hardest.}

We ran Annealed Importance Sampling (AIS) \cite{NealAIS} using
different versions of Hamiltonian Monte Carlo (HMC) methods for the transition kernel.
Such a highly correlated high-dimensional prior highlights some deficiencies in
a basic HMC method, where mixing can be slow due to a sample's leapfrog trajectory oscillating
up and down the sides of a valley of $\log \{ p(\y | \x)^{\beta} \prior(\x) \}$, without actually progressing
through it.
To further aid AIS with different HMC methods, we additionally let AIS anneal from
a Gaussian approximation $q(\x)$ to the GPC posterior, instead of the prior.
The approximation $q(\x)$ was obtained with Expectation Propagation (EP).

Figure \ref{fig:gpc} compares the estimates of $\log Z$ obtained with AIS to the required computation time.
The details of the methods are:

\begin{description}
\item[AIS$q$+HMC] in (g) runs AIS from the EP's $q(\x)$ at $\beta=0$ to $p(\x | \y)$
at $\beta = 1$ using intermediate distributions
\begin{equation} \label{eq:aisProb}
p_{\beta}(\x) = \frac{1}{Z(\beta)} \left( \prod_n \Phi(y_n x_n) \cdot {\cal N}(\x ; \0, \K) \right)^{\, \beta} q(\x)^{1 - \beta} \ .
\end{equation}
Note that a starred label indicates that the estimates were aided by $q(\x)$.
A HMC transition kernel with $l_{\max} = 200$ \emph{leapfrog} steps
is used at each $\beta \in [0,1]$ value.
AIS's $\beta$-grid is a geometric progression over $B = 400$ $\beta$-values.
Plot (g) used a step size $\epsilon = 0.02$ per proposal; both $l_{\max}$ and $\epsilon$ were carefully tuned to the problem.
The simplest AIS-HMC version, which anneals from $p(\x)$ and not $q(\x)$, didn't obtain estimates inside the bounds of Figure \ref{fig:gpc}, and is excluded. 
\item[AIS+RMHMC] in (j) and (l) anneals from $p(\x)$, and replaces HMC with a more advanced RMHMC
that uses $\epsilon = 0.1$ and $l_{\max} = 10$ leapfrog steps per proposal at each $\beta$ value.
\item[AIS$q$+RMHMC] in (c*), (f*), (g*) and (k*) anneals from $q(\x)$ using a RMHMC kernel
($\epsilon = 0.1, l_{\max} = 10$).
\end{description}

It is known that the EP estimate of $\log Z$ is remarkably accurate for this problem \cite{KussAssessing}, hence EP's $\log Z$ estimate and its a second-order corrected estimate
\cite{OpperPerturbative} are given for reference.


\bibliographystyle{elsarticle-num}
\bibliography{riemann}

\begin{thebibliography}{1}
\expandafter\ifx\csname url\endcsname\relax
  \def\url#1{\texttt{#1}}\fi
\expandafter\ifx\csname urlprefix\endcsname\relax\def\urlprefix{URL }\fi
\expandafter\ifx\csname href\endcsname\relax
  \def\href#1#2{#2} \def\path#1{#1}\fi

\bibitem{fraccaro16adaptive}
M.~Fraccaro, U.~Paquet, O.~Winther, An adaptive resample-move algorithm for
  estimating normalizing constants, arXiv:1604.01972.

\bibitem{RMHMC}
M.~Girolami, B.~Calderhead, Riemann manifold {L}angevin and {H}amiltonian
  {M}onte {C}arlo methods, Journal of the Royal Statistical Society, Series B
  73~(2) (2011) 123--214.

\bibitem{girolami09riemannian}
M.~Girolami, B.~Calderhead, S.~A. Chin, Riemannian manifold {H}amiltonian
  {M}onte {C}arlo, arXiv:0907.1100 (2009).

\bibitem{Neal2010}
R.~M. Neal, {MCMC} using {Hamiltonian} dynamics, Handbook of Markov Chain Monte
  Carlo 54 (2010) 113--162.

\bibitem{rasmussen06book}
C.~E. Rasmussen, C.~K.~I. Williams, Gaussian Processes for Machine Learning,
  MIT Press, 2006.

\bibitem{KussAssessing}
M.~Kuss, C.~E. Rasmussen, Assessing approximate inference for binary {G}aussian
  process classification, Journal of Machine Learning Research 6 (2005)
  1679--1704.

\bibitem{NealAIS}
R.~M. Neal, Annealed importance sampling, Statistics and Computing 11~(2)
  (2001) 125--139.

\bibitem{OpperPerturbative}
M.~Opper, U.~Paquet, O.~Winther, Perturbative corrections for approximate
  inference in {G}aussian latent variable models, Journal of Machine Learning
  Research 14~(Sep) (2013) 2857--2898.

\end{thebibliography}

\end{document}